\documentclass[journal]{IEEEtran}

\ifCLASSINFOpdf
\else
\fi

\ifCLASSOPTIONcompsoc
    \usepackage[caption=false, font=normalsize, labelfont=sf, textfont=sf]{subfig}
\else
	\usepackage[caption=false, font=footnotesize]{subfig}
\fi

\usepackage{cite}
\usepackage[version=4]{mhchem}
\usepackage{graphicx}
\usepackage{amsmath,amssymb,amsfonts}
\usepackage{color,soul}
\usepackage[table,xcdraw]{xcolor}
\usepackage{romannum}
\usepackage{graphicx}
\usepackage{commath}
\usepackage{textcomp}

\graphicspath{{figures/}}

\begin{document}

% \title{Memristive Spiking Neural Network with Unsupervised Learning} 
\title{SPICEprop: Backpropagating Errors Through Memristive Spiking Neural Networks}

\author{
Peng Zhou,~\IEEEmembership{Student Member,~IEEE,}
        Jason~K.~Eshraghian,~\IEEEmembership{Member,~IEEE,}\\
        Dong-Uk Choi,
        and Sung-Mo Kang,~\IEEEmembership{Life~Fellow,~IEEE}%<-this % stops a space
        
%\thanks{Manuscript received December 12, 2019.}
%\thanks{This work was supported by the NIH grant 5R01EB02390302}% <-this % stops a space
\thanks{P. Zhou, D. Choi, and S. M. Kang are with the Department of Electrical and Computer Engineering, UC, Santa Cruz, CA, USA.}%
\thanks{J. K. Eshraghian is with the Dept. of Electrical Engineering and Computer Science, University of Michigan, Ann Arbor, MI, USA.}%
}

\maketitle
\begin{abstract}
We present a fully memristive spiking neural network (MSNN) consisting of novel memristive neurons trained using the backpropagation through time (BPTT) learning rule. Gradient descent is applied directly to the memristive integrate-and-fire (MIF) neuron designed using analog SPICE circuit models, which generates distinct depolarization, hyperpolarization, and repolarization voltage waveforms. Synaptic weights are trained by BPTT using the membrane potential of the MIF neuron model and can be processed on memristive crossbars. The natural spiking dynamics of the MIF neuron model are fully differentiable, eliminating the need for gradient approximations that are prevalent in the spiking neural network literature. Despite the added complexity of training directly on SPICE circuit models, we achieve 97.58$\%$ accuracy on the MNIST testing dataset and 75.26$\%$ on the Fashion-MNIST testing dataset, the highest accuracies among all fully MSNNs.
%We have gained 100.0$\%$ accuracy for training and 99.5$\%$ accuracy for testing by using the MNIST dataset. 

\end{abstract}
% Note that keywords are not normally used for peer review papers.
\begin{IEEEkeywords}
Neuromorphic computing, memristor, spiking neural network, supervised learning, backpropagation.
\end{IEEEkeywords}

% \fi
\IEEEpeerreviewmaketitle
\vspace{-5pt}
\section{Introduction}
\IEEEPARstart{S}{piking} Neural Networks (SNNs), often referred to as the third generation of neural networks, draw inspiration from biological neurons to enable heightened efficiency over conventional neural networks.
%are considered more biologically plausible compared with other well-known Artificial Neural Networks (ANNs) such as Convolutional Neural Networks (CNNs) and Recurrent Neural Networks (RNNs). 
When processed on neuromorphic hardware, SNNs have shown significant energy and latency savings due to their activation sparsity, spike-based representations of data, and event-based data processing \cite{eshraghian2021training, azghadi2020hardware, zhou2022fully, zhou2020towards}. Recently, the remarkable success of backpropagation in the training of Deep Neural Networks (DNNs) has broadened and inspired research on SNNs. By unrolling the computational graph of spiking neuron models, SNNs can take advantage of the backpropagation through time (BPTT) algorithm \cite{werbos1990backpropagation} such that a global error is calculated in the final layer and backpropagated across the full history of the SNN dynamics. %via iterative application of the chain rule. 
By analogy, we propose that circuit-level neuron models described by dynamical state equations can also be trained in a similar manner. In doing so, the rich dynamics of nonlinear circuit elements, such as memristors, can be fully leveraged during the network optimization process.

% In addition, surrogate gradient descent \cite{neftci2019surrogate} can be applied to solve the non-differentiability of spike-generation due to hard thresholding of the membrane potential. This supervised learning paradigm has propelled SNN research forward, enabling it to become competitive with non-spiking networks after several decades of research. Several SNN Python packages have been developed to make these benefits more accessible to the broader research community \cite{snnTorch, sinabs, rockpool}.

The memristor, theoretically postulated by Chua in 1971 \cite{chua1971memristor} and generalized by Chua and Kang in 1976 \cite{chua1976memristive}, has become a commercially available technology that can be integrated in the back-end-of-the-line (BEOL) of modern CMOS processes \cite{eflash, Mad200}. The non-volatile retention capacity of the memristor is often likened to synaptic memory \cite{jo2010nanoscale, rahimi2020complementary, cai2019fully, serrano2013stdp}, and its threshold-switching characteristics have been exploited in spiking neuron models \cite{pickett2013scalable, eshraghian2018neuromorphic, zhang2017artificial, zhang2020brain}. %Its CMOS-compatibility, high-density, nanoscale vertical integration, and ability to directly implement biological features, as opposed to requiring several arithmetic steps as in transistor-only circuits have attracted much attention \cite{covi2016analog}. The memristor has been broadly investigated in the context of SNNs and integrated for neuromorphic computing. 
% However, much of the prior work on memristive spiking neural networks (MSNNs) is constrained to either memristive synapses \textit{or} memristive neurons, but not both. The work in \cite{wang2018fully} demonstrated a fully integrated memristive system that emulated both synapses and spiking neuron models to achieve simple pattern recognition tasks using in-house fabricated diffusive memristors, not readily accessible to the broader research community. 
However, much of the prior work on memristive spiking neural networks (MSNNs) is constrained to using either memristive synapses \textit{or} memristive neurons, but not both. Memristive synaptic arrays are used only for non-volatile memory retention, where networks are trained offline and thus, do not harness the complex dynamics of memristors. Networks of memristive neurons (`neuristors'), on the other hand, are challenging to train and only used in small-scale networks constrained to simple tasks, such as low-dimensional pattern recognition solved via associative or local learning rules. The work in \cite{wang2018fully} successfully integrates a fully memristive network, including memristive synapses and neurons, to perform basic pattern recognition, but the memristors used in this work are not accessible to the broader research community. The work presented in \cite{kiani2021fully} uses analog ReLU activations with memristive synapses to achieve 93.63$\%$ testing accuracy on the MNIST dataset. As a non-spiking network, analog signals are noise prone, whereas spikes are error tolerant discrete events distributed in time.

In this paper, we scale up the complexity of learnable tasks in fully memristive SNNs by directly applying gradient descent to the nonlinear state evolution of memristive neurons and synapses. Both neurons and synapses are modeled using memristors. The memristive integrate-and-fire (MIF) neuron model is designed to achieve distinct depolarization, hyperpolarization, and repolarization voltage phases with a minimal set of circuit elements. Memristive synapses interconnect layers of neurons.  %using quantized weight states. 
To train the fully memristive network, we propose SPICEprop, an application of error backpropagation to large-scale networks derived from SPICE circuit models. This enables dynamical, time-varying memristive neurons to achieve much higher accuracy on data-driven tasks than has been previously reported with MSNNs. By relying on the analog spiking characteristics that naturally occur in the MIF neuron model, the non-differentiability of spike-based activations are completely avoided. This means SPICEprop does not rely on surrogate gradient techniques that are commonly used to train SNNs, which calculates biased gradient estimators to circumvent the dead neuron problem \cite{neftci2019surrogate}. To promote broad accessibility of our methods, we use SPICE models of commercially available, low-cost memristors to demonstrate the efficacy of SPICEprop in a supervised deep learning framework.

% Although the training process utilizes the conventional hardware architectures such as CPU/GPU, the inference process with MIF neurons and quantized weights is shown to excel in view of power consumption and network density. With commercial fabrication of resistive random-access memories (RRAM) \cite{eflash, Mad200} using memristive cells, the memristor-based weight quantization enables a fully memristive neural network efficient for inference. We will show that a fully memristive neural network can be trained with BPTT to attain high accuracy. We use the SPICE models of commercially available, low-cost memristors \cite{molter2016generalized} to demonstrate the efficacy of MSNNs in supervised deep learning framework .
\vspace{-10pt}
\section{Methods}

\subsection{MIF model}
Our MSNN adopts the MIF neuron model shown in Fig.~\ref{mif2circuit} \cite{kang2021build}. The MIF neuron is characterized by the differential equations below:

\begin{figure}[htbp]
\centering
\includegraphics[scale=0.4]{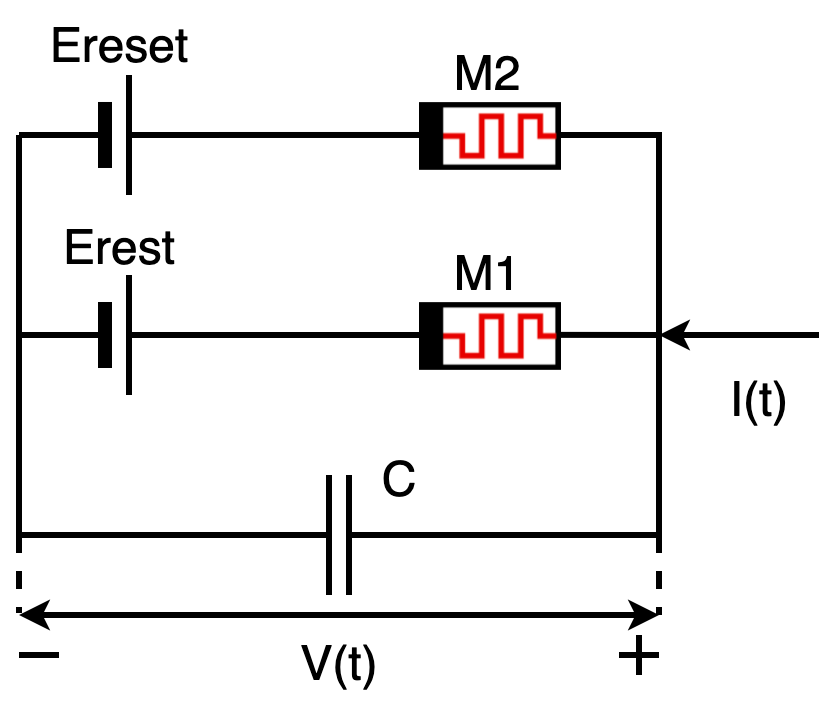}
\caption{A memristive integrate-and-fire (MIF) neuron consists of two memristors $M_1$ and $M_2$, connected to DC voltage sources $E_{\rm rest}$ and $E_{\rm reset}$, in parallel with a capacitor $C$. The MIF neuron is provably minimal in generating a membrane potential traversing from a rest voltage level to a threshold voltage level, then to a reset voltage level, and then back to the rest potential when a current pulse is applied.}
\label{mif2circuit}
\end{figure}

% \vspace{-8pt}
\begin{subequations}\label{mif2}
\begin{equation}
\frac{{\mathrm{d}}v}{\mathrm{d}t}=\frac{I - G_1(v-E_{\rm rest}) - G_2(v-E_{\rm reset})}{C} 
\end{equation}
\begin{equation}
\frac{{\mathrm{d}x_1}}{\mathrm{d}t}= \frac{1}{\tau_1}(\frac{1-x_1}{1+e^{\frac{v_{\rm on1}-(v-E_{\rm rest})}{k_{th}}}}  -   \frac{x_1}{1+e^{\frac{(v-E_{\rm rest})-v_{\rm off1}}{k_{th}}}}) 
\end{equation}
\begin{equation}
\frac{{\mathrm{d}x_2}}{\mathrm{d}t}= \frac{1}{\tau_2}(\frac{1-x_2}{1+e^{\frac{v_{\rm on2}-(v-E_{\rm reset})}{k_{th}}}}  -   \frac{x_2}{1+e^{\frac{(v-E_{\rm reset})-v_{\rm off2}}{k_{th}}}}) 
\end{equation}
\begin{equation}
G_1 = \frac{x_1}{R_{\rm on1}} + \frac{1-x_1}{R_{\rm off1}} 
\end{equation}
\begin{equation}
G_2 = \frac{x_1}{R_{\rm on2}} + \frac{1-x_2}{R_{\rm off2}} 
\end{equation}
\end{subequations}

\noindent where $G_1$ and $G_2$ are the memductances, $x_1$ and $x_2$ are a pair of internal states, $\tau_1$ and $\tau_2$ are time constants governing the rate of change in internal states, and $k_{th}$ is the effective thermal voltage. They are the characteristic variables of $M_1$ and $M_2$, respectively. This system of equations mirrors several prominent SPICE memristor models, and has been used to emulate the commercially available KNOWM\textsuperscript{TM} memristor.

For a physically plausible system, discrete spikes must be represented as continuous current inputs. An input spike train of Dirac delta pulses $\sum_n{\delta(t-t_j^n)}$ is used to generate an alpha-shaped input current $I$, modeled by Eq.~(2):
\begin{subequations}\label{alphaCurrent}
\begin{align}
\tau_{syn} \frac{{\mathrm{d}I}}{\mathrm{d}t} &= a-I \\
\tau_{syn} \frac{{\mathrm{d}a}}{\mathrm{d}t} &= -a + W_i \cdot \sum_n{\delta(t-t_i^n)}
\end{align}
\end{subequations}

\noindent where $a$ is an internal state variable, $\tau_{syn}$ is a time constant governing the shape of the alpha current, $W_i$ is the synaptic weight between presynaptic neuron $i$ and associated postsynaptic neuron. %$\sum_n{\delta(t-t_j^n)}$ represents the spike train applied to the presynaptic neuron $j$, which ultimately generates $I$ is the input current.
% The dynamics of the alpha synapse and MIF neuron with the resultant memristor internal states and voltage waveforms are shown in Fig.~\ref{mif2result}.

% \begin{figure}[htbp]
%   \begin{center}
%   \includegraphics[width=3in]{figure/time.png}
%   \caption{Simulation results of an alpha shaped input current applied to a MIF neuron. Top: Alpha synapse dynamics. Middle: Internal MIF states $x_1$, $x_2$. Bottom: Voltage response $v$. }\label{mif2result}
%   \end{center}
% \end{figure}

The parameters used in this network are listed in Table \ref{tablemsnn1}.

\begin{table}[htb]
\caption{MIF circuit parameters}
\label{tablemsnn1}
\begin{center}
\begin{tabular}{|c|c|c|c|}
\hline
\textbf{Parameter} & \textbf{Value} & \textbf{Parameter} & \textbf{Value} \\ \hline 
$E_{\rm rest}$ & 0 mV & $E_{\rm reset}$ & 50 mV \\
$C$ & 100 pF & $k_{th}$ & 15 mV \\
$v_{\rm off_1}$, $v_{\rm off_2}$ & 5 mV & $v_{\rm on_1}$, $v_{\rm on_2}$ & 110 mV \\
$R_{\rm off_1}$, $R_{\rm off_2}$ & 0.1 M$\Omega$ & $R_{\rm on_1}$, $R_{\rm on_2}$ & 1 k$\Omega$ \\
$\tau_1$, $\tau_2$ & 100 steps & $\tau_{syn}$ & 100 steps \\
\hline
\end{tabular}
\end{center}
\vspace{-9pt}
\footnotesize{}
\end{table}

\subsection{Numerical Integration of MIF Model}
The SPICE model of the memristor used in the MIF neuron described in Eq.~(1) must be adapted to a recurrent form to be compatible in a deep learning framework. SPICE simulators employ a variety of ordinary differential equation (ODE) solvers, which, in the first order, are often equivalent to the backward Euler method. For compatibility with BPTT, we adopt the forward Euler method to find an approximate solution to the MIF network, as the forward Euler method provides an explicit update equation. As such, we retain the temporal dynamics of the MIF neuron, and also derive a discrete time representation of the MIF neuron that can be unrolled across time steps for use with the BPTT algorithm.
%We use the ordinary differential equations (ODEs) of the MIF neuron in Eq.~(1). In order to use the MIF neuron in a deep learning framework, similar to \cite{neftci2019surrogate}, a computational graph of an MSNN in discrete time is needed, and a numerical integration version of the MIF neuron based on Eq.~(1) is also required. In numerical analysis, the Runge–Kutta methods are most popular. Among them, the Euler method is the simplest and its computation efficiency is the highest. 
The approximate solution of a MIF neuron model with alpha current dynamics using the forward Euler method is provided in Eq.~(3).
%We use a forward Euler version of MIF incorporated with alpha current dynamics as described in Eq.~(3).

\begin{subequations}\label{euler}
\begin{equation}
a[t+1] = -\frac{a[t]}{\tau_{syn}} + a[t] +  \sum_j{W_j \cdot S_{j}}
\end{equation}
\begin{equation}
I[t+1] = \frac{a[t]-I[t]}{\tau_{syn}} + I[t]
\end{equation}
\begin{gather*}
x_1[t+1] = \frac{1}{\tau_1} \Big(\frac{1-x_1[t]}{1+e^{\frac{v_{\rm on1}-(v[t]-E_{\rm rest})} {k_{th}}}}  - 
\end{gather*}
\begin{gather}
\frac{x_1[t]}{1+e^{ \frac{(v[t]-E_{\rm rest})-v_{\rm off1}}{k_{th}}}} \Big)+ x_1[t]
\end{gather}
\begin{gather*}
x_2[t+1] = \frac{1}{\tau_2} \Big(\frac{1-x_2[t]}{1+e^{\frac{v_{\rm on2}-(v[t]-E_{\rm reset})} {k_{th}}}}  - 
\end{gather*}
\begin{gather}
\frac{x_2[t]}{1+e^{ \frac{(v[t]-E_{\rm reset})-v_{\rm off2}}{k_{th}}}} \Big)+ x_2[t]
\end{gather}
\begin{equation}
G_1[t+1] = \frac{x_1[t+1]}{R_{\rm on1}} + \frac{1-x_1[t+1]}{R_{\rm off1}}
\end{equation}
\begin{equation}
G_2[t+1] = \frac{x_2[t+1]}{R_{\rm on2}} + \frac{1-x_2[t+1]}{R_{\rm off2}}
\end{equation}
\begin{gather*}
v[t+1] = \frac{I[t]-G_1[t](v[t]-E_{\rm rest})-G_2[t](v[t]-E_{\rm reset})}{C}
\end{gather*}
\begin{gather}
+ v[t]
\end{gather}
% \begin{gather}
% + v[t]
% \end{gather}
\end{subequations}

\begin{figure}[htbp]
  \begin{center}
  \includegraphics[width=3in]{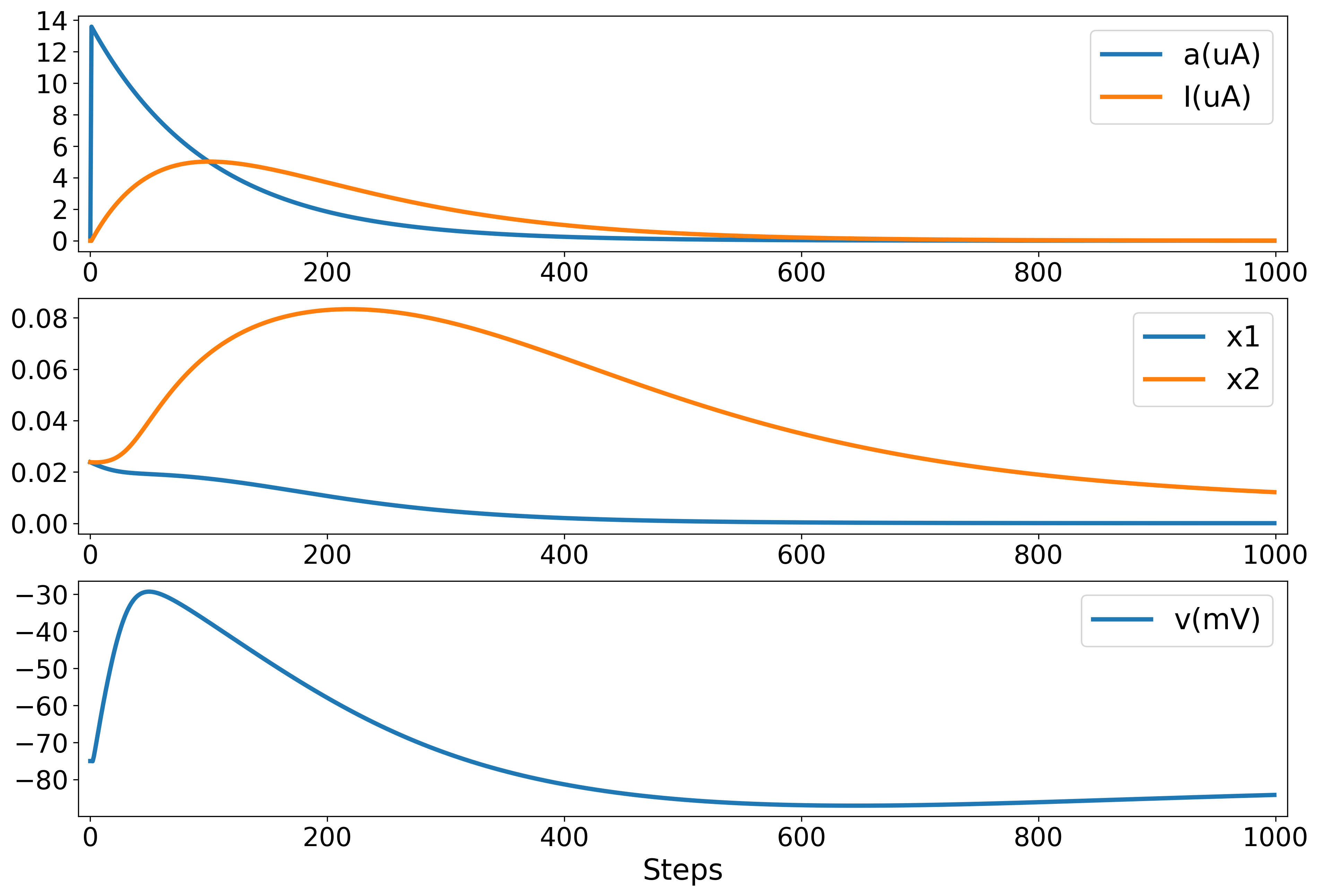}
  \caption{Simulation results of MIF neuron solved using forward Euler numerical integration. The results matches the quantitative dynamics of solved by a SPICE simulator. Top: Alpha synaptic current dynamics. Middle: Internal states $x_1$, $x_2$. Bottom: Voltage response $v$.}\label{mif2Euler}
  \end{center}
  \vspace{-10pt}
\end{figure}

Fig.~\ref{mif2Euler} shows the simulation results across 1,000 time steps. Each time step is of duration 0.01 ms and the total time simulated is 10 ms. The rest, integrate, threshold, and reset performances are correctly simulated by a SPICE simulator to match the dynamics of the alpha synapse current and the MIF neuron modeled by the ODEs in Eq.~(1) and (2).

\subsection{Backpropagation Through Time in Memristive Spiking Neural Networks}
\begin{figure}[htbp]
  \begin{center}
  \includegraphics[width=3in]{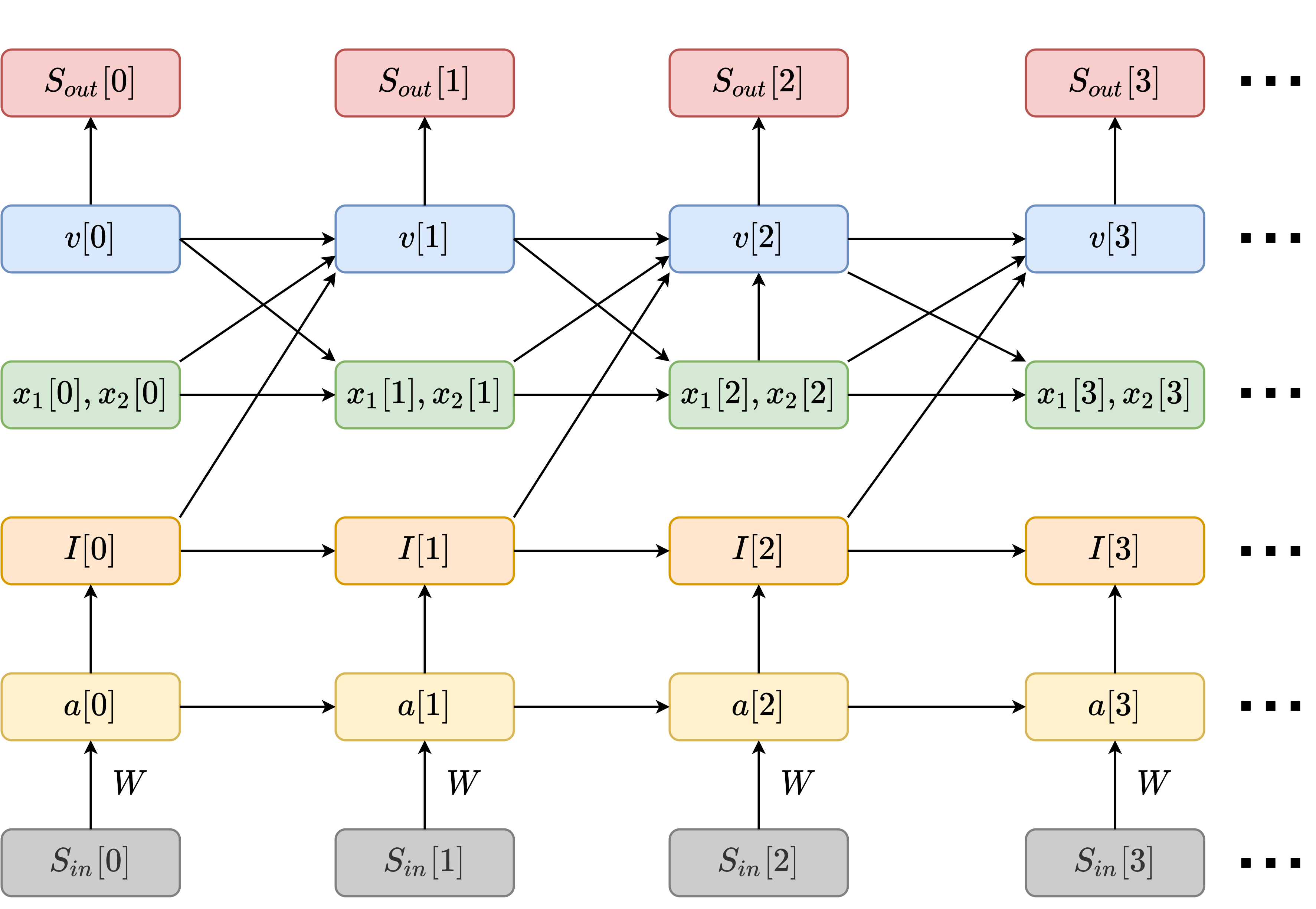}
  \caption{Unrolled MSNN computational graph in discrete time.}\label{snnrnn}
  \end{center}
  \vspace{-10pt}
\end{figure}

The discrete-time solution in Eq.~(3) is illustrated as an MSNN computational graph in Fig.~\ref{snnrnn}, resembling an unrolled recurrent neural network. The loss can be calculated based on either spiking activity or the membrane potential. Two observations support the use of the membrane potential. Firstly, an increase in potential of the correct output neuron drives the neuron to spike, while decreasing the potential of the incorrect neuron classes makes these neurons silent. Applying the target to the membrane potential can serve as a proxy for modulating the spiking behavior. Secondly, the MIF membrane potential is autonomously reset after a spike is elicited without the need for extra circuitry; i.e., the spike is part of the membrane potential waveform which is weighted and propagated to downstream layers. Therefore, a fully analog MSNN can be trained without reliance on additional surrogate gradient calculation steps, as used in almost all multi-layer SNNs trained via error backpropagation \cite{neftci2019surrogate, eshraghian2021training}. 

\subsection{Network Architecture}
\begin{figure}[htbp]
  \begin{center}
  \includegraphics[width=3.2in]{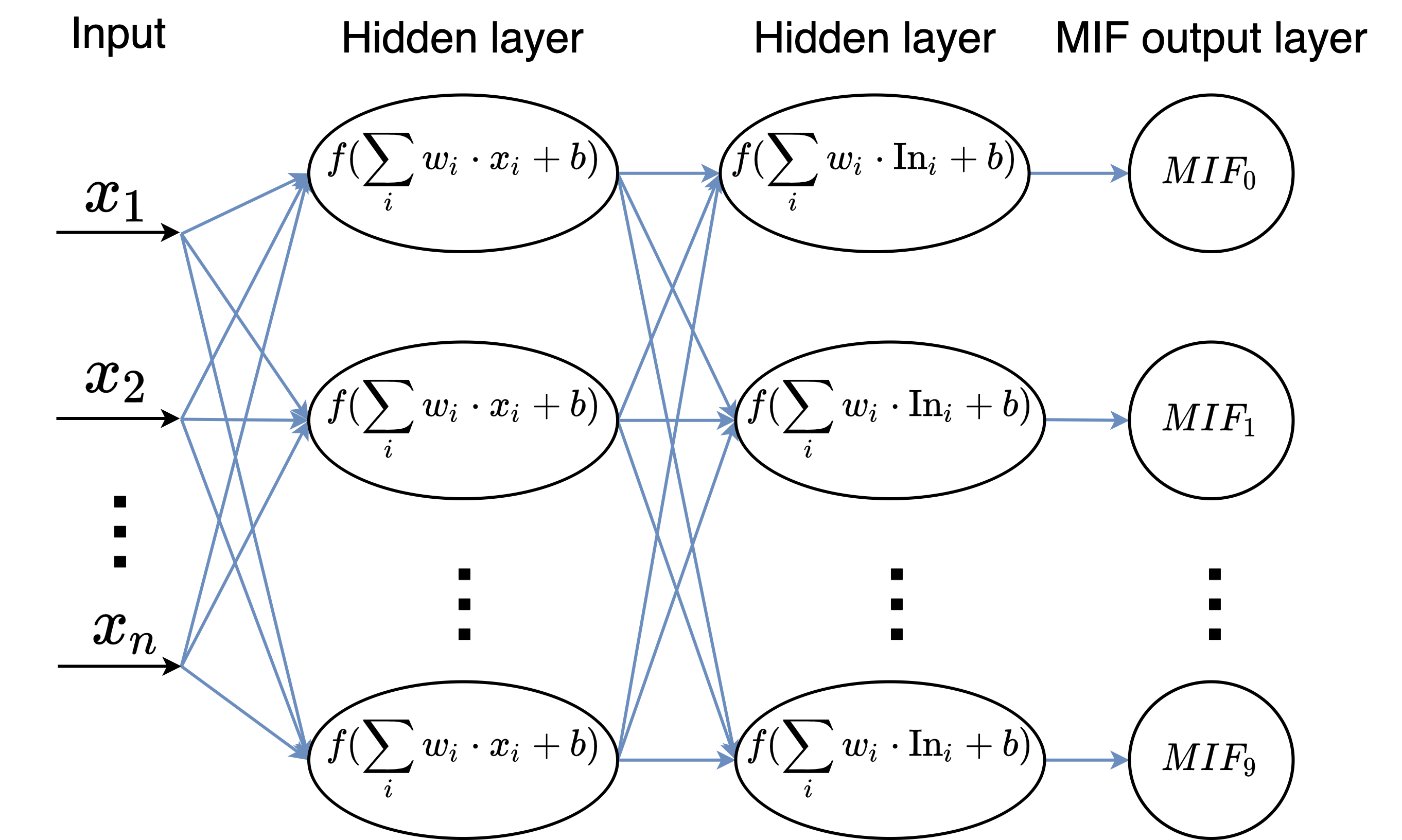}
  \caption{The MSNN architecture.}\label{mifdl}
  \end{center}
  \vspace{-10pt}
\end{figure}

To validate the use of MIF neurons in a deep learning framework, we used a 4-layer neural network shown in Fig.~\ref{mifdl}. In the input layer, there are $28\times28=784$ input units. The first hidden layer consists of 100 neurons followed by a ReLU activation, and the second hidden layer contains 10 neurons followed by a ReLU activation. Finally, the output layer consists of 10 MIF neurons. Fully connected (all-to-all) weights are used between each layer.

\section{Results}
\begin{figure}[!htbp]
\centering
\subfloat[]
{
	\includegraphics[width=3in]{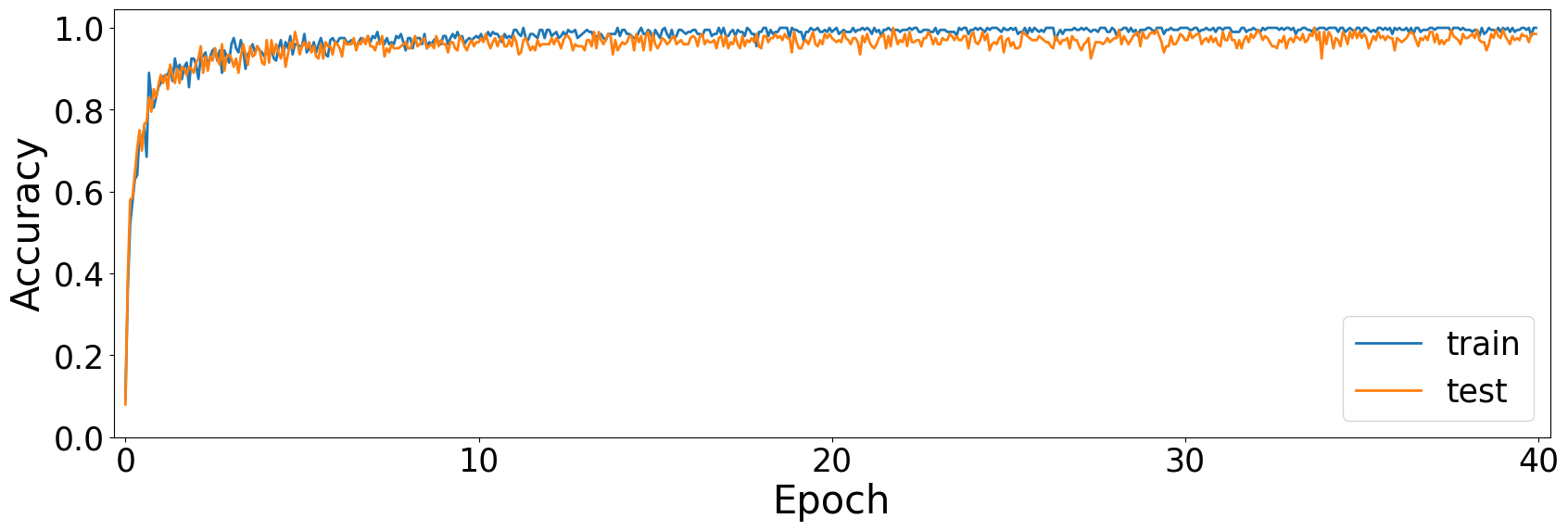}
	\label{mnist}
}\\
\subfloat[]
{
	\includegraphics[width=3in]{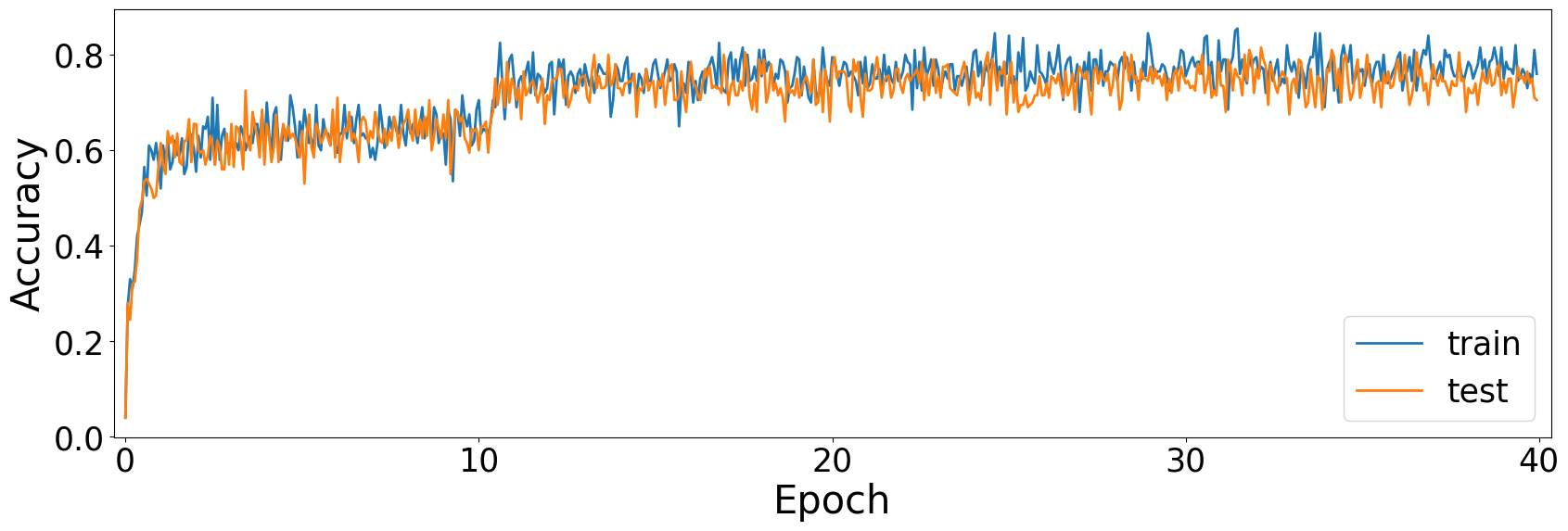}
	\label{fmnist}
}
\caption{Accuracy across epochs for training and testing processes for (a) MNIST dataset (b) Fashion-MNIST dataset.}\label{result_curve}
\vspace{-5pt}
\end{figure}

The automatic differentiation engine from PyTorch v1.10.1 was used to perform BPTT through the network in Python 3.8. For training and testing, we used the MNIST dataset \cite{lecun1998gradient}  which consists of 70,000 samples of handwritten digits. During the training process, 1,000 time steps are simulated for each input image sample. Over the 1,000 steps, the input pixel intensity is fed to the network only at steps 0, 400, and 800 to promote sparse network activity. The negative log-likelihood loss is applied to the membrane potential, i.e., $v(t)$ in the MIF circuit in Fig.~\ref{mif2circuit}, at each time step. In other words, the goal of the network is to maximize the voltage across the MSNN circuit of the correct class. The total loss is summed prior to backpropagating the error through the SPICE memristor model. The Adam optimizer is utilized as it performs well on both recurrent networks and stochastic problems \cite{kingma2014adam}. 
For the Adam optimizer, the learning rate is set to $10^{-4}$, the initial decay rates for first and second order moments are set to $\beta_1=0.9$, $\beta_2=0.999$. 
The learnable weights of dense layers are initialized by sampling from a uniform distribution $U(-\sqrt k, \sqrt k)$, 
where $k = \frac{1}{N_i}$ and $N_i$ is the number of input features to layer $i$. The MSNN is trained for 40 epochs using a batch size of 200 samples. Accuracy is evaluated at every 10 training iterations. 
%During the training, we found the ReLU layers are required in order to avoid unstable performance of forward Euler method. 
For the MNIST dataset, %we show that the MSNN is capable of achieving 100.0$\%$ accuracy on a batch of the training set and 99.5$\%$ accuracy on a batch of the test set. 
the training and testing accuracies across multiple epochs are shown in Fig.~\ref{result_curve}. %: the accuracy reaches close to 100$\%$ after approximately 5 epochs and remains stable for both datasets. 
97.58$\%$ accuracy is achieved for the total MNIST test set. For a more challenging task using real-world data, we also train our MSNN on the Fashion-MNIST dataset, where  %both training and testing batch accuracies achieves close to 80$\%$ after approximately 10 epochs and remains stable. 
we obtain 75.26$\%$ testing accuracy. This demonstrates the first functional result on the Fashion-MNIST dataset using analog MSNNs, and highlights the potential for MSNNs to be capable of moving beyond simple MNIST classification.

\section{Discussion and Conclusion}
We have confirmed the compatibility of circuit-level MSNNs with error-driven supervised learning. Our primary results show that MIF neurons based on SPICE models can be trained in a multi-layer neural network using BPTT, side-stepping the need for surrogate gradient signals. To the best of our knowledge, this is the first work to perform gradient descent directly on a network of SPICE models of memristive neurons that emit analog spikes. 

While MSNNs are extremely power efficient, they are also highly challenging to train. We achieve state of the art accuracy on MNIST and Fashion-MNIST datasets when compared to all other networks of spiking MSNNs. Prior attempts on the MNIST dataset using memristive spiking neurons with error-driven optimization \cite{duan2020spiking} only achieved 83.24$\%$. The approach in \cite{duan2020spiking} aims to fit memristive neurons to LIF neuron models, whereas our approach directly relies on the gradient of the naturally arising memristive dynamics during training, in order to achieve the accuracy boost to 97.58$\%$. 

Although SNNs add more latency for training due to the additional temporal dimension, there are many reasons why this approach is valuable and holds great potential:
\begin{itemize}
% \item SNNs are naturally compatible with neuromorphic computing chips.
\item SNNs can exploit sparse data connections and activations to reduce memory access frequency, which we exploit by only presenting input data for 3 out of the total 1,000 time steps of simulation. %Additionally, neuromorphic chips usually take less bits for synaptic connection compared with CPU/GPU which typically use high precision values. %For weight quantization trained with BPTT, memristors are well suited in a neuromorphic chip, consuming much less power than high precision networks 
\cite{schuman2017survey, wang2020integration, zhang2020brain, camunas2019neuromorphic}.
\item A large-scale neural network based on the MIF neuron has a much higher density compared with the conventional CMOS counterparts by taking advantage of vertical BEOL CMOS integration.
\item Area and power can be saved as less peripheral circuits are needed by eliminating the need for analog-to-digital converters (ADCs) and digital-to-analog converters (DACs) in bit-line current summation approaches; i.e., all spiking is a result of the natural evolution of the memristor's state dynamics.
\item The natural temporal dynamics of the MIF model allow for differentiable spike propagation through a MSNN, eliminating the need for biased gradient estimators such as surrogate gradients.
\end{itemize}

Future work will involve further increasing the complexity of the datasets, accounting for the non-idealities present in memristive synapses, and stochasticity that is not designed in the SPICE model yet.

% Our future work will involve testing different datasets and exploring more network architectures. More work is required to simulate multiple MIF layers in the MSNNs using other numerical integration methods, especially to overcome the shortcomings of the forward Euler method.

% \section{Conclusion}

% or


% \section*{Acknowledgment}

\ifCLASSOPTIONcaptionsoff
  \newpage
\fi

\bibliographystyle{IEEEtran}
\bibliography{references}


% \begin{IEEEbiography}{Sung Mo Kang}
% \end{IEEEbiography}

%\begin{IEEEbiographynophoto}{Peng Zhou}


%\vfill


\end{document}